\documentclass[letterpaper, 10 pt, conference]{ieeeconf}  

\IEEEoverridecommandlockouts                              

\usepackage{amsmath} 
\usepackage{amssymb}  
\usepackage{graphicx}
\usepackage{algorithm} 
\usepackage{algpseudocode} 

\title{\LARGE \bf
Goal-Driven Autonomous Exploration Through Deep Reinforcement Learning}

\author{Reinis Cimurs$^{1}$, Il Hong Suh$^{2}$, and Jin Han Lee$^{2}$
\thanks{*This work was supported by the Technology Innovation Program (Industrial Strategic Technology Development, 10080638) funded by the Ministry of Trade, Industry \& Energy (MOTIE), Republic of Korea.}

\thanks{$^{1}$R. Cimurs is with Industry-University Cooperation Foundation, Hanyang University, Republic of Korea;
        {\tt\small reinis@incorl.hanyang.ac.kr}}%
\thanks{$^{2}$J.H. Lee, I.H. Suh are with Cogaplex company, Republic of Korea. J.H. Lee is also with Hanyang University
        {\tt\small \{jhlee,ihsuh\}@cogaplex.com}}%

}

\begin{document}

\maketitle
\thispagestyle{empty}
\pagestyle{empty}

\begin{abstract}

In this paper, we present an autonomous navigation system for goal-driven exploration of unknown environments through deep reinforcement learning (DRL).
Points of interest (POI) for possible navigation directions are obtained from the environment and an optimal waypoint is selected, based on the available data.
Following the waypoints, the robot is guided towards the global goal and the local optimum problem of reactive navigation is mitigated.
Then, a motion policy for local navigation is learned through a DRL framework in a simulation.
We develop a navigation system where this learned policy is integrated into a motion planning stack as the local navigation layer to move the robot between waypoints towards a global goal.
The fully autonomous navigation is performed without any prior knowledge while a map is recorded as the robot moves through the environment.
Experiments show that the proposed method has an advantage over similar exploration methods, without reliance on a map or prior information in complex static as well as dynamic environments.

\end{abstract}

\section{INTRODUCTION}
Over the last couple of decades, the field of simultaneous localization and mapping (SLAM) has been studied extensively.
Typically, in SLAM systems a person operates a measuring device and a map is generated from the location and the environmental landmarks \cite{yousif2015overview}.
The operator makes decisions, which parts of the previously unmapped environment to visit and their visitation order.
More so if the requirement is to map a path between two locations.
Humans can use their best knowledge of their surroundings and instincts to locate possible pathways to the goal, even if working in unknown environments.
Afterward, they can manually guide the mapping robot along the selected path to the global goal.
However, it is not always possible to manually operate the mapping device due to various reasons - the high cost of labor, physical constraints, limited resources, environmental dangers, and others.
Subsequently, autonomous exploration is an area that is garnering a lot of attention with ongoing research in the ability to delegate navigation and mapping tasks to autonomous vehicles.
But unlike regular environment exploration, where the task is only to map the surroundings, fully autonomous goal-driven exploration is a twofold problem.
First, the exploration robot needs to make a decision on where to go to have the highest possibility of arriving at the global goal.
Without prior information or the global goal in sight, the system needs to point to possible navigation directions directly from the sensor data.
From such points of interest (POI) the best possible one needs to be selected as a waypoint to guide the robot to the global goal in the most optimal way.
Second, a robot motion policy, that does not depend on map data, for uncertain environments needs to be obtained.
With the recent advances in deep reinforcement learning (DRL) for robot navigation, high-precision decision-making has become feasible for autonomous agents.
Using DRL, an agent control policy can be learned to achieve the target task in an unknown environment \cite{sugiyama2015statistical}.
However, due to its reactive nature and lack of global information, it easily encounters the local optimum problem, especially for large-scale navigation tasks \cite{aberdeen2003policy}.

Therefore, in this paper, we present a fully autonomous exploration system for navigation to a global goal, without the necessity of human control or prior information about the environment.
Points of interest are extracted from the immediate vicinity of the robot, evaluated, and one of them is selected as a waypoint.
Waypoints guide the DRL-based motion policy to the global goal, mitigating the local optimum problem. 
Then, motion is performed based on the policy, without requiring fully mapped representation of the surroundings.
The main contributions of this work can be itemized as follows:
\begin{itemize}
    \item Designed a global navigation and waypoint selection strategy for goal-driven exploration.
    \item Developed a TD3 architecture based neural network for mobile robot navigation.
    \item Combined the DRL motion policy with global navigation strategy to mitigate the local optimum problem for exploration of unknown environments and performed extensive experiments to validate the proposed method.
\end{itemize}

\begin{figure}
    \centering
    \includegraphics[width=\columnwidth]{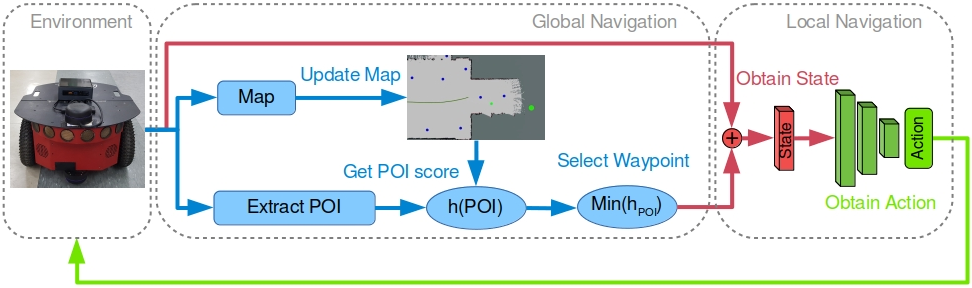}
    \vspace{-0.7cm}
    \caption{Navigation system implementation. Robot setup is depicted on the left. Global and local navigation with their individual parts and data flow is visualized in the middle and right, respectively.}
    \label{fig:robot}
\vspace{-0.65cm}
\end{figure}

The remainder of this paper is organized as follows.
In Section \ref{WORKS} related works are reviewed.
The proposed exploration system is described in Section \ref{NAVIGATION}.
Experimental results are given in Section \ref{EXPERIMENTS}.
Finally, this paper concludes in Section \ref{CONCLUSIONS}.

\section{RELATED WORKS} \label{WORKS}

Environmental exploration and mapping has been a popular study in the field of robotics for decades \cite{zelinsky1992mobile, surmann2003autonomous, chen2019learning}.
With the availability of various low-cost sensors and computational devices, it has become possible to perform SLAM on the robotic agent in real-time with a number of different approaches.
Sensor devices such as cameras \cite{mur2017orb, yu2018ds, cui2019sof}, two dimensional \cite{li2020gp, jiang2019fft, krinkin2018evaluation} and three dimensional LiDARs \cite{ren2019robust, pierzchala2018mapping, li2018efficient}, and their combinations \cite{liang2016visual,chan2018robust, zhang2018pirvs} are used not only to detect and record the environment but also to autonomously position the agents within it.
However, to obtain reliable map information of the surroundings through navigation, a large portion of developed SLAM systems rely on human operators or a pre-described plan \cite{taketomi2017visual, filipenko2018comparison, 7759248}.
Also, many autonomous exploration approaches are developed based on previously available map \cite{von2017monocular, gao2018autonomous, palomeras2019active}.
POI for exploration are extracted from the map edges that show free space in the environment.
Subsequently, a path is planned and navigation performed towards the selected POI \cite{keidar2014efficient, gao2018improved, tang2019autonomous}.

Due to the rise in popularity and the capabilities of deep learning methods, reliable robot navigation through neural networks has been developed.
Robots are able to perform deliberate motions obtained directly from neural network outputs. 
In \cite{shrestha2019learned} the environment is mapped by using learned predictions of the frontier areas that guide the exploration of the robot.
But the navigation is still carried out based on a planner.
Exploratory navigation through deep Q-learning is presented in \cite{tai2016robot}, where a robot learns to avoid obstacles in unknown environments.
This is further extended in \cite{xie2017towards}, where a robot learns an obstacle-avoiding policy in a simulation and the network is applied to real-world environments.
Here, discrete navigation actions are selected to avoid obstacles without a specific goal.
Robot actions in continuous action space are obtained by performing learning in deep deterministic policy gradient-based networks.
In \cite{tai2017virtual, cimurs2020goal} the environment state information is combined with the goal position to form the input for the network.
However, since the information about the environment is given locally in time and space, these networks may encounter a local optimum problem.
Even though these methods provide high accuracy and reliability in their respective domains, they are difficult to deploy in practice due to concerns relating to safety, generalization ability, local optimum problem, and others.
As such, the neural network-based methods can tackle modular tasks, but might not be suitable for implementation in global, end-to-end solutions.
In \cite{chen2017socially} the learned, behavior-based navigation is successfully combined with planning.
However, local planners are still used for motion planning and the neural network is used only to avoid dynamic obstacles in a previously mapped environment.
Similarly, in \cite{pokle2019deep, faust2018prm} creation of a local plan is learned and combined with a global path planner to avoid obstacles along the way.
Here, a local path is calculated through a neural network in an already mapped environment and exploration is not considered.

Therefore, we propose combining a lightweight learned motion policy with a broader global navigation strategy to solve a goal-driven exploration problem.
The navigation system's aim is not only to navigate around obstacles but also to explore and map out an unknown environment towards a specified global goal.
The proposed fully autonomous system is presented in Fig. \ref{fig:robot}.

\begin{figure}
    \centering
    \includegraphics[width=\columnwidth]{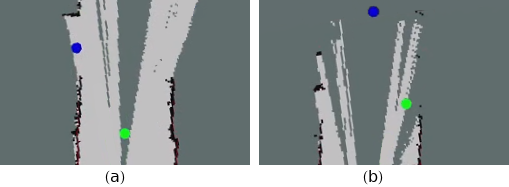}
    \vspace{-0.65cm}
    \caption{POI extraction from the environment. Blue circles represent an extracted POI with the respective method. Green circles represent the current waypoint. (a) Blue POI is obtained from a gap between laser readings. (b) Blue POI extracted from non-numerical laser readings. }
    \label{fig:nodes}
    \vspace{-0.65cm}
\end{figure}
\section{GOAL-DRIVEN AUTONOMOUS EXPLORATION} \label{NAVIGATION}

To achieve autonomous navigation and exploration in an unknown environment, we propose a navigation structure that consists of two parts: global navigation with optimal waypoint selection from POI and mapping; a deep reinforcement learning-based local navigation.
Points of interest are extracted from the environment and an optimal waypoint is selected following the evaluation criteria.
At every step, a waypoint is given to the neural network in the form of polar coordinates, relative to the robot's location and heading.
An action is calculated based on the sensor data and executed towards the waypoint.
Mapping is performed while moving between waypoints towards the global goal.

\begin{figure*}
    \centering
    \includegraphics[width=\textwidth]{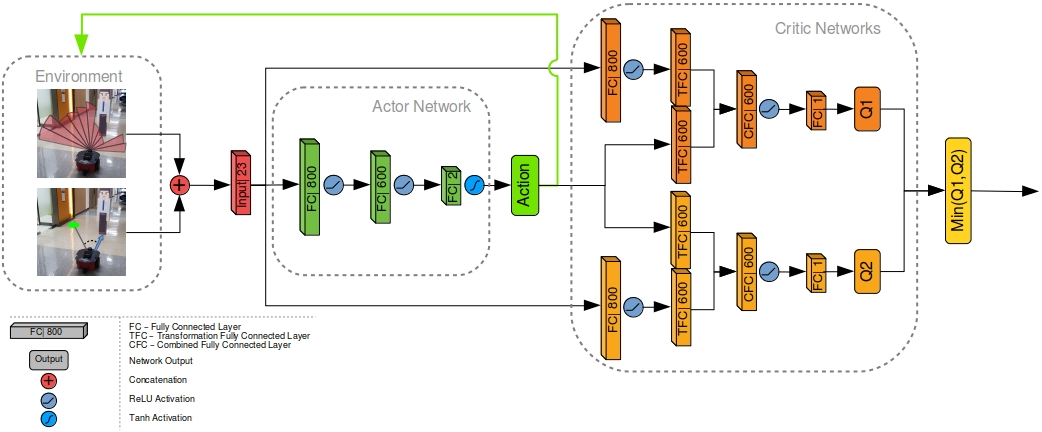}
    \vspace{-0.8cm}
    \caption{TD3 network structure including the actor and critic parts. Layer type and the number of their respective parameters are described within the layers. TFC layers refer to transformation fully connected layers $\tau$ and CFC layer to combined fully connected layer $L_c$.}
    \label{fig:network}
    \vspace{-0.6cm}
\end{figure*}
\subsection{Global Navigation}

For the robot to navigate towards the global goal, intermediate waypoints for local navigation need to be selected from available POI.
As there is no initial information about the environment, it is not possible to calculate an optimal path.
Therefore, the robot needs not only to be guided towards the destination but also to explore the environment along the way to recognize possible alternative routes if it were to encounter a dead-end.
Since no prior information is given, the possible POI need to be obtained from the immediate surroundings of the robot and stored in the memory.

We implement two methods of obtaining new POI:
\begin{itemize}
    \item A POI is added if a value difference between two sequential laser readings is larger than a threshold, allowing the robot to navigate through the presumed gap.
    \item Due to laser sensors having maximum range, the readings outside of this range are returned as a non-numerical type and represent free space in the environment. A POI is placed in the environment if sequential laser readings return a non-numerical value.
\end{itemize}
Examples of POI extraction from the environment are depicted in Fig. \ref{fig:nodes}.

If in subsequent steps any of the POI are found to be located near an obstacle, they are deleted from the memory.
A POI will not be obtained from the laser readings in a place that the robot has already visited.
Additionally, if a POI is selected as a waypoint but cannot be reached over a number of steps, it is deleted and a new waypoint selected.


From available POI, the optimal waypoint at the time-step $t$ is selected by using the Information-based Distance Limited Exploration (IDLE) evaluation method \cite{cimurs2021information}.
The IDLE method evaluates the fitness of each candidate POI as:
\begin{equation} \label{eq: heur}
    h(c_i) = tanh\left ( \frac{e^{\left ( \frac{d(p_t,c_i)}{l_2-l_1}\right )^2}}{e^{\left ( \frac{l_2}{l_2-l_1}\right )^2}} \right )l_2 + d(c_i,g) + e^{I_{i,t}},
\end{equation}
where score $h$ of each candidate POI $c$ with index $i$ is a sum of three components.
The Euclidean distance component $d(p_t,c_i)$ between robots position $p$ at $t$ and candidate POI is expressed as a hyperbolic tangent $tanh$ function:
\begin{equation} \label{eq: tanh}
tanh\left ( \frac{e^{\left ( \frac{d(p_t,c_i)}{l_2-l_1}\right )^2}}{e^{\left ( \frac{l_2}{l_2-l_1}\right )^2}} \right )l_2,
\end{equation}
where $e$ is the Euler's number, $l_1$ and $l_2$ are two-step distance limits in which to discount the score.
The two-step distance limits are set based on the area size of the DRL training environment.
The second component $d(c_i,g)$ represents Euclidean distance between the candidate and the global goal $g$.
Finally, map information score at $t$ is expressed as:
\begin{equation} \label{eq: map}
    e^{I_{i,t}},
\end{equation}
where $I_{i,t}$ is calculated as:
\begin{equation}
  I_{i,t} =   \frac{\sum_{w = -\frac{k}{2}}^{\frac{k}{2}}\sum_{h = -\frac{k}{2}}^{\frac{k}{2}}c_{(x+w)(y+h)}}{k^2}.
\end{equation}
$k$ is size of the kernel to calculate the information around the candidate points coordinates $x$ and $y$, $w$ and $h$ represent the kernel width and height, respectively.


A POI with the smallest IDLE score from (\ref{eq: heur}) is selected as the optimal waypoint for local navigation.

\subsection{Local Navigation} 
In a planning-based navigation stack, local motion is performed following the local planner.
In our approach, we replace this layer with a neural network motion policy.
We employ DRL to train the local navigation policy separately in a simulated environment. 

A Twin Delayed Deep Deterministic Policy Gradient (TD3) based neural network architecture is used to train the motion policy \cite{fujimoto2018addressing}.
TD3 is an actor-critic network that allows performing actions in continuous action space.
The local environment is described by bagged laser readings in 180$^{\circ}$ range in front of the robot \cite{choi2020fast}.
This information is combined with polar coordinates of the waypoint with respect to the robot's position.
The combined data is used as an input state $s$ in the actor-network of the TD3.
The actor-network consists of two fully connected (FC) layers.
Rectified linear unit (ReLU) activation follows after each of these layers.
The last layer is then connected to the output layer with two action parameters $a$ that represent the linear velocity $a_1$ and angular velocity $a_2$ of the robot.
A $tanh$ activation function is applied to the output layer to limit it in the range $(-1,1)$.
Before applying the action in the environment, it is scaled by the maximum linear velocity $v_{max}$ and the maximum angular velocity $\omega_{max}$ as follows:
\begin{equation}
a = \left [v_{max}\left(\frac{a_1+1}{2}\right), \omega_{max}a_2\right ].
\end{equation}
Since the laser readings only record data in front of the robot, motion backward is not considered and the linear velocity is adjusted to only be positive. 

The $Q$ value of the state-action pair $Q(s,a)$ is evaluated by two critic-networks.
Both critic-networks have the same structure but their parameter updates are delayed allowing for divergence in parameter values.
The critic-networks use a pair of the state $s$ and action $a$ as an input.
The state $s$ is fed into a fully connected layer followed by ReLU activation with output $L_{s}$.
The output of this layer, as well as the action, are fed into two separate transformation fully connected layers (TFC) of the same size $\tau_1$ and $\tau_2$, respectively.
These layers are then combined as follows:
\begin{equation}
    L_{c} = L_{s}W_{\tau_1} + a W_{\tau_2} + b_{\tau_2},
\end{equation}
where  $L_{c}$ is the combined fully connected layer (CFC), $W_{\tau_1}$ and $W_{\tau_2}$ are the weights of the $\tau_{1}$ and $\tau_{2}$, respectively. 
$b_{\tau_2}$ is bias of layer $\tau_{2}$.
Then ReLU activation is applied to the combined layer.
Afterward, it is connected to the output with 1 parameter representing the $Q$ value.
The minimum $Q$ value of both critic-networks is selected as the final critic output to limit the overestimation of the state-action pair value.
The full network architecture is visualized in Fig. \ref{fig:network}.

The policy is rewarded according to the following function:
\begin{equation}
r(s_t,a_t)   =\begin{cases}
r_g & \text{ if } D_t<\eta_D \\ 
r_c  & \text{ if collision} \\ 
v - |\omega| & \text{ otherwise},
\end{cases}
\end{equation}
where the reward $r$ of the state-action pair $(s_t,a_t)$ at timestep $t$ is dependant on three conditions.
If the distance to the goal at the current timestep $D_t$ is less than the threshold $\eta_D$, a positive goal reward $r_g$ is applied.
If a collision is detected, a negative collision reward $r_c$ is applied.
If both of these conditions are not present, an immediate reward is applied based on the current linear velocity $v$ and angular velocity $\omega$.
To guide the navigation policy towards the given goal, a delayed attributed reward method is employed following calculation:
\begin{equation}
 r_{t-i} = r(s_{t-i},a_{t-i})+\frac{r_g}{i},\: \: \: \: \: \: \: \:  \forall i= \left \{ 1,2,3,...,n \right\}  ,
\end{equation}
where $n$ is the number of previous steps where rewards are updated.
This means that the positive goal reward is attributed not only to the state-action pair at which the goal was reached, but also decreasingly over the last $n$ steps before it.
The network learned a local navigation policy that is capable of arriving at a local goal, while simultaneously avoiding obstacles directly from the laser inputs.

\subsection{Exploration and Mapping}
Following the waypoints, the robot is guided towards the global goal.
Once the robot is near the global goal, it navigates to it.
The environment is explored and mapped along the way.
Mapping uses laser and robot odometry sensors as sources and obtains an occupancy grid map of the environment.
The pseudo-code of the fully autonomous exploration algorithm with mapping is described in Algorithm 1.


    


\begin{algorithm}
\caption{Goal-Driven Autonomous Exploration}
\begin{algorithmic}[1]

    \State $globalGoal$ \Comment{Set global goal}
    \State $\delta$ \Comment{Set threshold of navigating to global goal}
    \While{$reachedGlobalGoal \not= True$} 
            \State Read sensor data
            \State Update map from sensor data
            \State Obtain new POI
            \If{$D_t<\eta_D$}
                \If {$waypoint = globalGoal$}
                    \State $reachedGlobalGoal = True$
                \Else
                    \If {$d(p_t,g)<\delta$} 
                        \State $waypoint \leftarrow\ globalGoal$
                    \Else
                        \For{$i$ in $POI$} 
                            \State calculate $h(i)$ from (\ref{eq: heur})
                        \EndFor
                        \State $waypoint \leftarrow\ POI_{min(h)}$
                    \EndIf
                \EndIf
    
            \EndIf
            \State Obtain an action from TD3
            \State Perform action
    \EndWhile  

\end{algorithmic}
\end{algorithm}

\section{EXPERIMENTS} \label{EXPERIMENTS}

Experiments in real-life settings of varying complexity were executed to validate the proposed goal-driven exploration system.

\subsection{System Setup}
\begin{figure}
    \centering
    \includegraphics[width=\columnwidth]{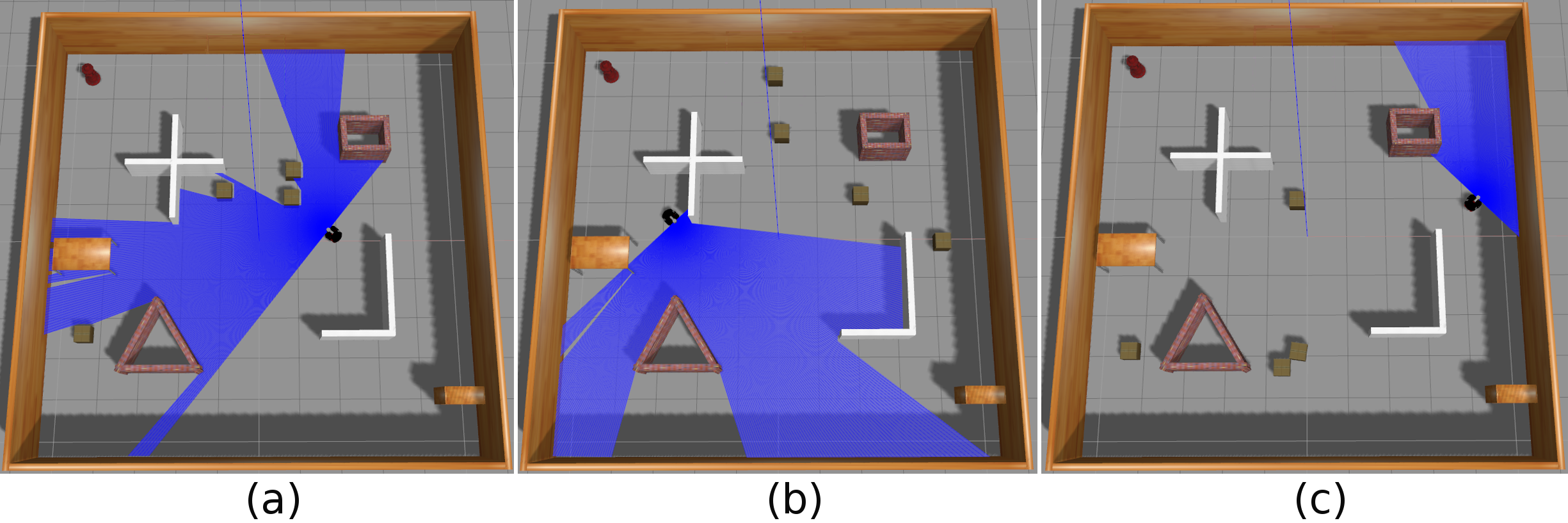}
    \vspace{-0.65cm}
    \caption{Examples of training environments. Blue field represents the input laser readings and range. The four box obstacles change their location on every episode as presented in (a), (b), and (c) in order to randomize the training data. }
    \label{fig:envs}
   \vspace{-0.65cm}
\end{figure}

The learning of local navigation through DRL was performed on a computer equipped with an NVIDIA GTX 1080 graphics card, 32 GB of RAM, and Intel Core i7-6800K CPU.
The TD3 network was trained in the Gazebo simulator and controlled by the Robot Operating System (ROS) commands.
The training ran for 800 episodes which took approximately 8 hours.
Each training episode concluded when a goal was reached, a collision was detected or 500 steps were taken.
$v_{max}$ and $\omega_{max}$ were set as 0.5 meters per second and 1 radian per second, respectively.
The delayed rewards were updated over the last $n=10$ steps and parameter update delay was set as 2 episodes.
The training was carried out in a simulated 10x10 meter-sized environment depicted in Fig \ref{fig:envs}. 
Gaussian noise was added to the sensor and action values to facilitate generalization and policy exploration.
To create a varied environment, the locations of the box-shaped obstacles were randomized at the start of each episode.
Example of their changing locations is depicted in Fig. \ref{fig:envs}(a),(b) and (c).
The robots starting position and the goal locations were randomized on every episode.
\begin{table*}[]
\caption{Results of the First Quantitative Experiment}
\vspace{-0.3cm}
\label{table: exp1}
\begin{center}
\begin{tabular}{|l|c|c|c|c|c|c|c|c|c|c|}
\hline
               & \textbf{Av. Dist.(m)} & \textbf{Av. T.(s)} & \textbf{Min. Dist.(m)} & \textbf{Min. T.(s)} & \multicolumn{1}{l|}{\textbf{Max. Dist.(m)}} & \multicolumn{1}{l|}{\textbf{Max. T.(s)}} & \textbf{$\sigma$(Dist.)} & \multicolumn{1}{l|}{\textbf{$\sigma$(T)}} & \multicolumn{1}{l|}{\textbf{Av. Map(m$^2$)}} & \multicolumn{1}{l|}{\textbf{Goals}} \\ \hline
\textbf{GD-RL} & 77.02             & 171.82            & 70.17              & 147.07             & 79.49                                    & 184.98                                   & 3.86          & 14.96                              & 580.63                                   & 5/5                                 \\ \hline
\textbf{NF}    & 40.74             & 109.11            & 34.67              & 74.6               & 50.5                                     & 150.25                                   & 6.73          & 29.31                              & 475.75                                   & 5/5                                 \\ \hline
\textbf{LP-AE} & 40.02             & 123.56            & 35.43              & 96.48              & 46.02                                    & 155.4                                    & 4.95          & 25.49                              & 491.57                                   & 5/5                                 \\ \hline
\textbf{GDAE}  & 41.42             & 88.03             & 34.77              & 68.24              & 54.06                                    & 118.05                                   & 8.33          & 20.42                              & 492.01                                   & 5/5                                 \\ \hline
\textbf{PP}    & 32.26             & 61.53             & 32.03              & 59.52              & 32.32                                    & 63.88                                    & 0.22          & 1.89                               & -                                        & 5/5                                 \\ \hline
\end{tabular}
\vspace{-0.65cm}
\end{center}
\end{table*}

For quantitative experiments and comparison to similar methods with constrained resources, the network was transferred to an Intel NUC mini-PC (2.70 GHz i7-8559U CPU, 16GB of RAM) that facilitated the full exploration system.
For qualitative experiments, the system was embedded on a laptop with an NVIDIA RTX 2070M graphics card, 16 GB of RAM, and Intel Core i9-10980HK CPU running Ubuntu 18.04 operating system.
ROS Melodic version managed the packages and control commands.
Pioneer P3-DX mobile platform was used as the robot base.

To ensure safe navigation, the robot was equipped with two RpLidar laser sensors at different heights with a maximal measuring distance of 10 meters.
Robot setup is displayed in Fig. \ref{fig:robot}.
The location and angle of both lasers were calibrated and laser readings were recorded in 180$^{\circ}$ in front of the robot.
The data from each device was bagged into 21 groups, where the minimum value of each group was selected as the representative sensor value.
The minimal value of each respective bag was selected to create the final laser input state of 21 values.
The final laser data was then combined with the polar coordinates to the waypoint. 
The mapping of the environment was performed based on the full laser readings of the top RpLidar sensor in combination with the internal mobile robot odometry.
ROS package SLAM Toolbox \cite{macenskislam} was used to obtain and update the global map of the environment as well as localize the robot within it.
Maximal linear and angular velocities were set to the same values as in the simulation.
Kernel size $k$ was set as 1.5 meters and $l_1$, $l_2$ values in (\ref{eq: tanh}) were selected as 5 and 10 meters, respectively.
The waypoints and global goal were considered reached at a 1-meter distance.

\subsection{Quantitative Experiments}

In order to quantify-ably evaluate the proposed method, it was compared to different environment exploration methods in indoor settings.
We refer to the proposed method as Goal-Driven Autonomous Exploration (GDAE) which combines reinforcement learning with global navigation strategy to arrive at the global goal.
To the best of the author's knowledge, currently, there are no comparable goal-driven exploration methods that employ neural network-based motion policies.
Therefore, a state-of-the-art Nearest Frontier (NF) exploration strategy from \cite{da2020novel} was employed for comparison with a planning-based method.
Here, the distance factor for the nearest frontier was updated to include the distance to the goal.
Additionally, GDAE was compared to a reinforcement learning-based method without the global navigation strategy, referred to as Goal-Driven Reinforcement Learning (GD-RL).
Experiments with the proposed navigation system were carried out, where the neural network was substituted with a ROS local planner package (TrajectoryPlanner).
We refer to this system as Local Planner Autonomous Exploration (LP-AE).
Finally, as a baseline comparison, a path obtained with the Dijkstra algorithm in an already known map was executed. 
We refer to this method as Path Planner (PP).
Experiments were performed five times with each method in two environments.
The recorded data includes travel distance (Dist.) in meters, travel time (T) in seconds, recorded map size in square meters, and how many times has the method successfully reached the goal.
Average (Av.), maximal (max.), minimal (min.), and standard deviation ($\sigma$) were calculated from the obtained results.
The recorded map size was calculated only from known pixels.

\begin{figure}
    \centering
    \includegraphics[width=\columnwidth]{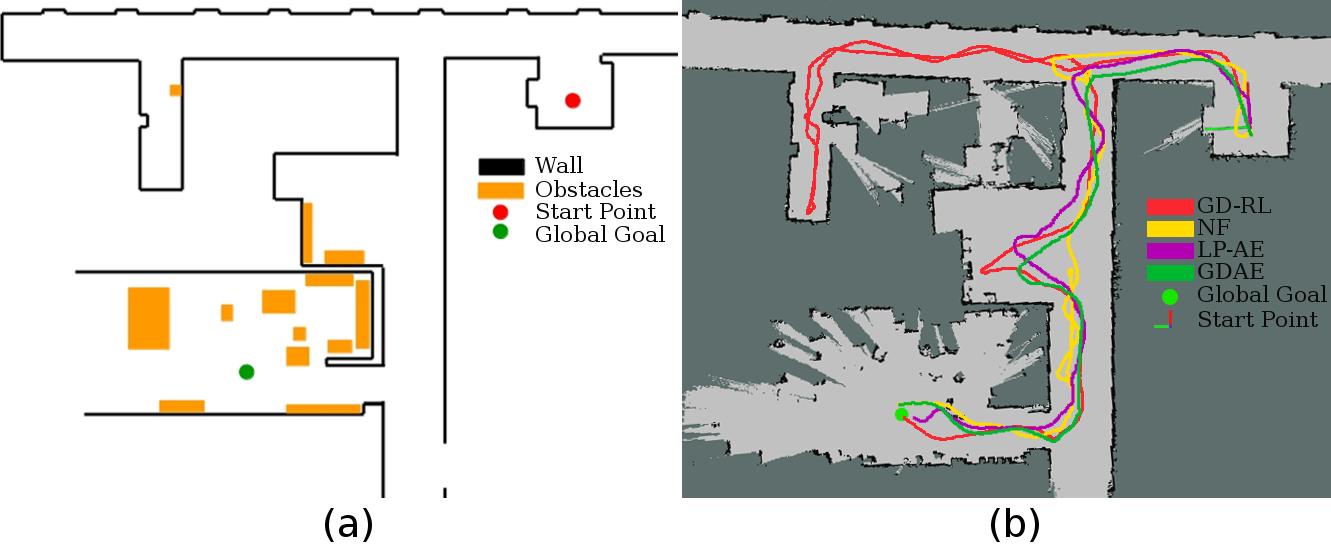}
    \vspace{-0.65cm}
    \caption{Environment and resulting navigation examples of the respective methods in the first quantitative experiment. (a) Depiction of the experiment environment. (b) Example of a resulting map of the environment with overlaid path of each method. }
    \label{fig:exp1}
    \vspace{-0.7cm}
\end{figure}

The first environment is depicted in Fig. \ref{fig:exp1} and consisted mostly of smooth walls with multiple local optima.
The goal was located at coordinate (-12,15).
While all the methods were able to arrive at the global goal, they did so with differing travel distance and time.
GDAE was capable of arriving at the global goal at comparable travel distance to similar methods but the navigation took less time.
When planning-based methods (NF and LP-AE) obtained a new waypoint, a new path needed to be calculated.
With constrained resources, the robot needed to wait until it received the new path. 
On the contrary, the proposed method started navigation towards the new waypoint instantly.
This was more evident for LP-AE, where each time a waypoint was selected, a new local trajectory calculation was required.
However, the GD-RL method fell in the local optimum trap, from which it escaped by following a wall.
This significantly increased the distance and time to the global goal.
The results of the experiment are described in Table \ref{table: exp1}.
\begin{table*}[]
\caption{Results of the Second Quantitative Experiment}
\vspace{-0.5cm}
\label{table: exp2}
\begin{center}
\begin{tabular}{|l|c|c|c|c|c|c|c|c|c|c|}
\hline
               & \textbf{Av. Dist.(m)} & \textbf{Av. T.(s)} & \textbf{Min. Dist.(m)} & \textbf{Min. T.(s)} & \multicolumn{1}{l|}{\textbf{Max. Dist.(m)}} & \multicolumn{1}{l|}{\textbf{Max. T.(s)}} & \textbf{$\sigma$(Dist.)} & \multicolumn{1}{l|}{\textbf{$\sigma$(T)}} & \multicolumn{1}{l|}{\textbf{Av. Map(m$^2$)}} & \multicolumn{1}{l|}{\textbf{Goals}} \\ \hline
\textbf{GD-RL} & 73.12             & 206.02            & 68.97              & 172.95             & 77.37                                    & 234.12                                   & 4.2           & 30.88                              & 986.92                                   & $3^*$/5                                \\ \hline
\textbf{NF}    & 56.44             & 188.02            & 52.81              & 178.9              & 63.13                                    & 205.78                                   & 5.79          & 15.37                              & 1014.87                                  & 3/5                                 \\ \hline
\textbf{LP-AE} & 82.59             & 296.68            & 56.13              & 187.16             & 103.89                                   & 431.27                                   & 19.73         & 100.76                             & 1031.25                                  & 4/5                                 \\ \hline
\textbf{GDAE}  & 67.79             & 156.54            & 54.31              & 124.3              & 79.69                                    & 197.0                                    & 9.66          & 27.64                              & 791.6                                    & 5/5                                 \\ \hline
\textbf{PP}    & 50.53             & 81.16             & 48.7               & 79.34              & 53.32                                    & 83.15                                    & 2.04          & 1.6                                & -                                        & 5/5                                 \\ \hline
\end{tabular}
\end{center}
* Successful only after human interference.
\vspace{-0.3cm}
\end{table*}

\begin{figure*}
    \centering
    \includegraphics[width=\textwidth]{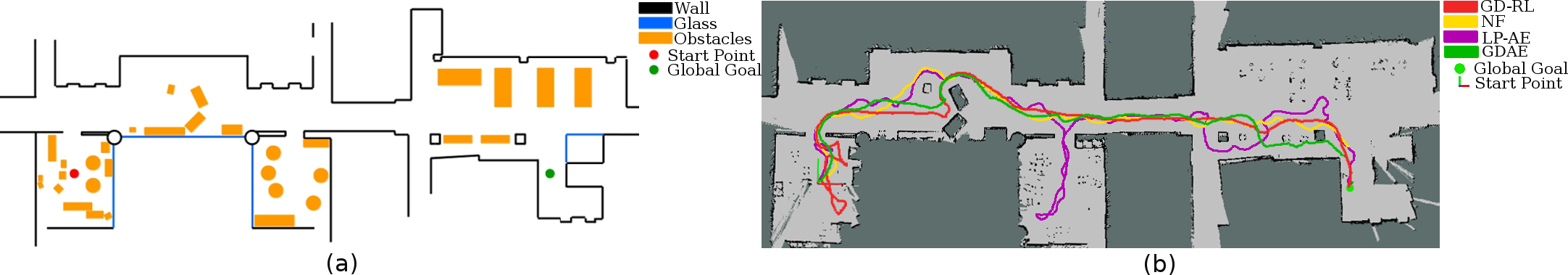}
    \vspace{-0.7cm}
    \caption{Environment with complex obstacles and resulting navigation examples of the respective methods in the second quantitative experiment. (a) Depiction of the experiment environment. (b) Example of a resulting map of the environment with overlaid path of each method. }
    \label{fig:exp2}
    \vspace{-0.5cm}
\end{figure*}

The second experiment is depicted in Fig. \ref{fig:exp2} and introduced obstacles of various complexity into the environment, such as - furniture, chairs, tables, glass walls, and others.
The start point was located in a local optimum area with a see-through glass wall at height of the top laser.
The proposed method successfully and reliably navigated to the global goal in the shortest time and comparable distance.
The NF method calculated the path through the glass wall and tried to navigate through it on two out of the five runs eventually failing to reach the global goal. 
Similarly, the LP-AE collided with an obstacle by creating a path through the legs of a chair.
These obstacles were initially detected as sensor noise, thus required multiple readings to be confirmed and recorded as an obstacle in the map.
During the first two runs, the GD-RL method was unable to escape the local optimum and failed to reach the global goal.
In subsequent runs, a human operator intervened as the robot approached the escape from the starting area and guided the robot towards the free space.
The obtained results were calculated over the successful runs and are described in Table \ref{table: exp2}.

\subsection{Qualitative Experiments}
\begin{figure}
    \centering
    \includegraphics[width=0.8\columnwidth]{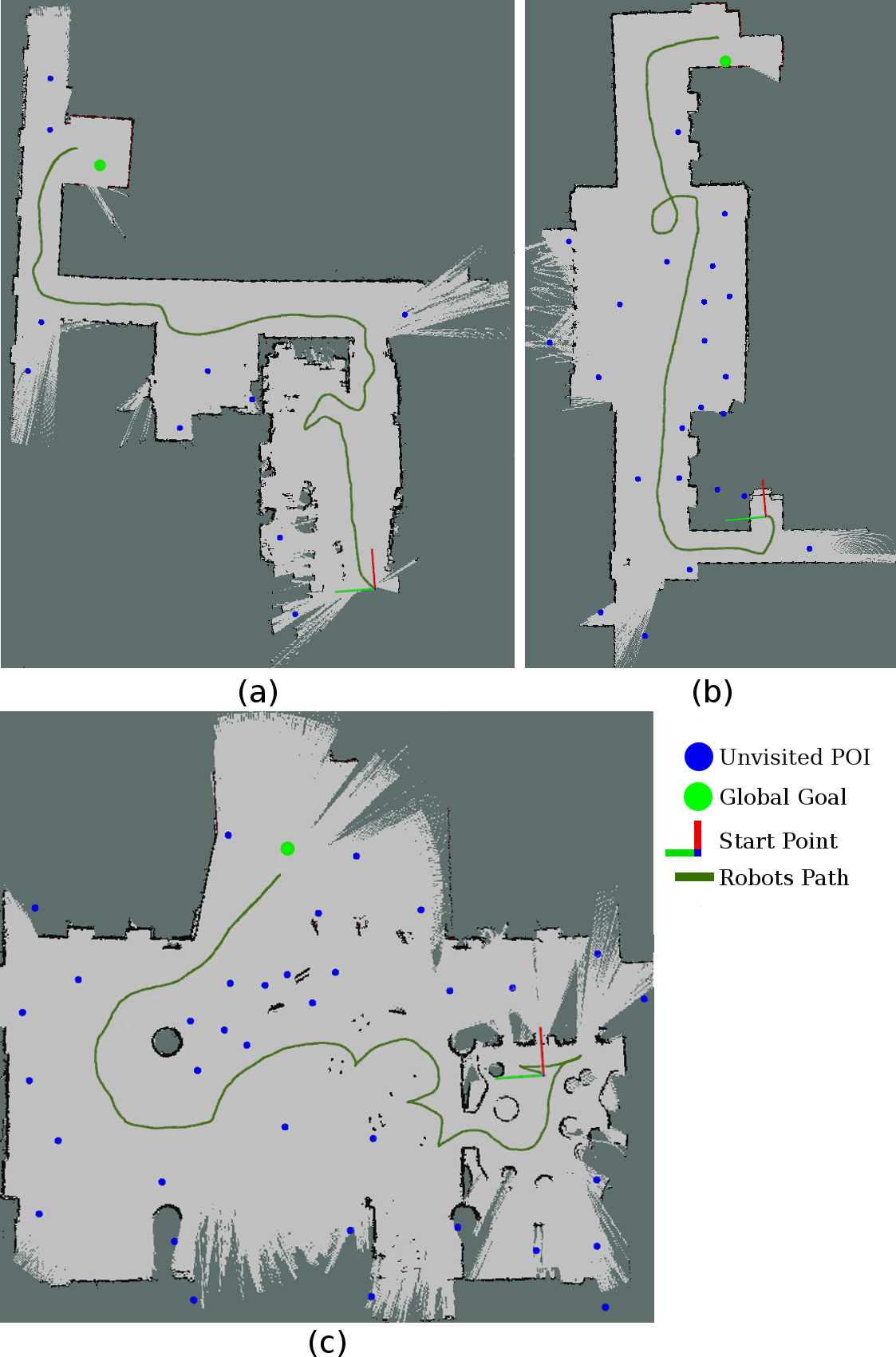}
    \vspace{-0.3cm}
    \caption{Experimental results of goal driven exploration and mapping. (a) Navigation towards a goal in hallway. (b) Navigation towards a goal in a hallway, when starting position is in a local optimum. (c) Navigation towards a goal in a hall through a cluttered environment when starting position is in a local optimum. }
    \label{fig:local}
    \vspace{-0.8cm}
\end{figure}

Additional experiments with GDAE method were performed in various indoor settings.
In Fig. \ref{fig:local} three different environments are depicted, where a local optimum needed to be avoided or navigated out of to arrive at the global goal.
The blue dots represent the obtained POI, green path represents the robot's path.
\begin{figure*}
    \centering
    \includegraphics[width=0.8\textwidth]{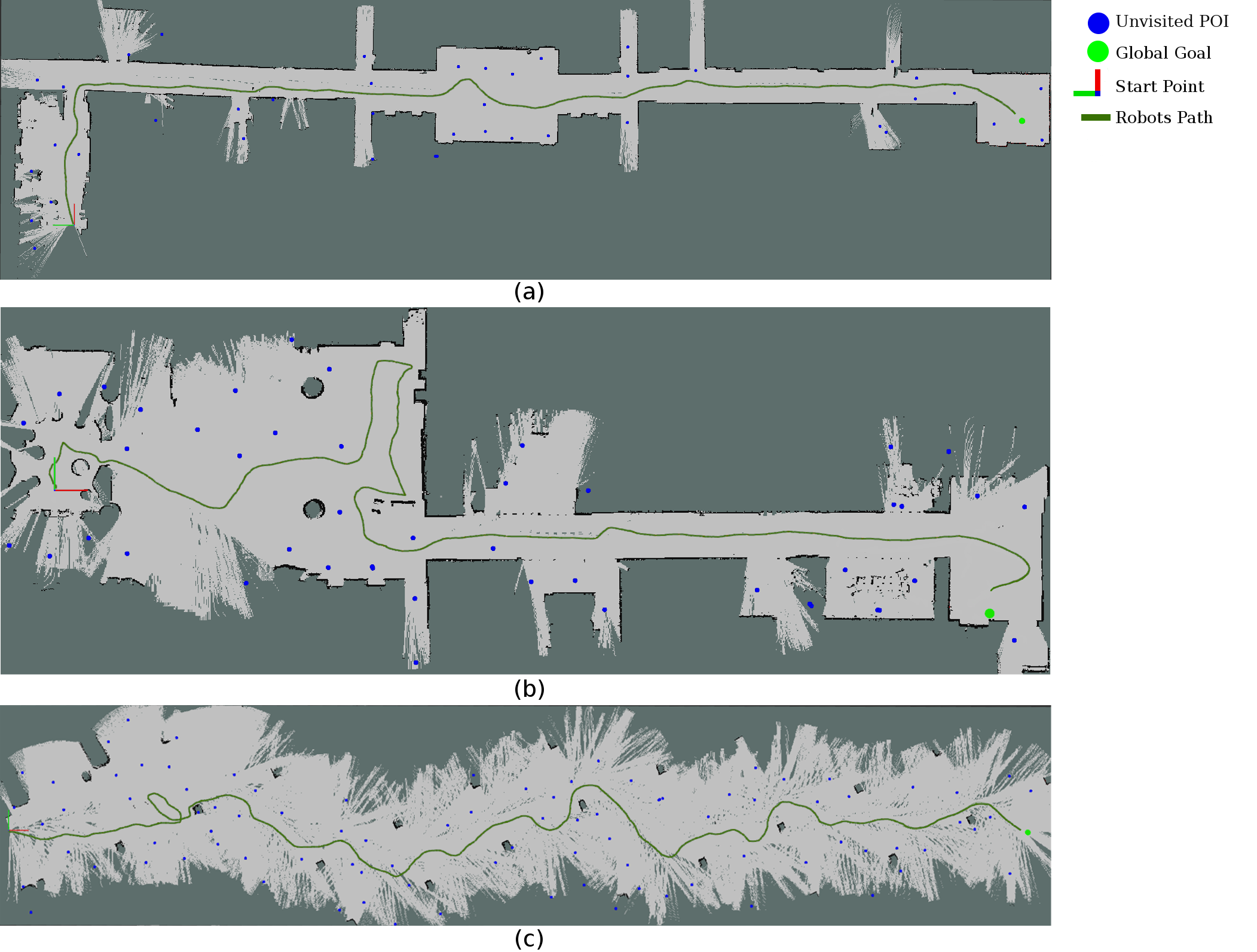}
    \vspace{-0.35cm}
    \caption{Experimental results of goal driven exploration and mapping with goal located at a significant distance from the robot. (a) Navigation towards a goal in hallway. (b) Navigation towards a goal in a hall and hallway with backtracking. (c) Navigation towards a goal in an underground parking lot with an incomplete map information. }
    \label{fig:global}
    \vspace{-0.7cm}
\end{figure*}
In Fig. \ref{fig:local}(a) the goal position was located at coordinate (22, 12).
A straight path towards the goal led towards a room corner and a local optimum.
Once the robot had obtained the information about a corner, it was able to backtrack, find the exit of the room, and navigate towards the global goal through the hallways.
In Fig. \ref{fig:local}(b) the goal position was located at coordinate (23, 2), and the robot starting location was placed in the local optimum.
Without prior map information and previously extracted POI, the robot did not have sufficient information for backtracking.
By using the global navigation strategy, the local environment was explored, which allowed the robot to navigate out of the confined hallway and to the global goal.
In Fig. \ref{fig:local}(c) the goal coordinates were (10, 10) and the robot starting location was the local optimum.
The environment was littered with obstacles of various shapes and sizes, such as chairs, signs, potted plants, and others.
The robot successfully explored its surroundings, navigated through the clutter using the local navigation, and arrived at the global goal.

Afterward, navigation and mapping experiments were performed on a larger scale.
In Fig. \ref{fig:global} three scenarios are introduced, where a robot navigated towards a goal in hallways and an underground parking lot.
In Fig. \ref{fig:global}(a) goal was located at coordinate (12, -90).
The robot followed the selected waypoints out of the room and navigated towards the global goal.
In Fig. \ref{fig:global}(b) goal coordinate was (60, -7).
The robot's starting position was located in a cluttered environment and its starting pose was compromised by facing an obstacle.
The robot was capable of navigating out of the clutter and moved towards the global goal.
Once it approached the wall that blocked direct movement towards the global goal, it explored the local surroundings to find a path.
In Fig. \ref{fig:global} the goal was located 100 meters diagonally across an underground parking lot at coordinate (100, 0).
The environment was generally free of major obstacles and walls that would appear on a map.
However, there was a multitude of parking bumpers at a maximum height of approximately 0.2 meters.
Since the mapping algorithm took only the top laser sensor as the input, the parking bumpers were not mapped and a plan could not be constructed to navigate around them.
The local reactive navigation used both sensors as inputs and was capable of detecting the bumpers.
Therefore, the robot was successful in navigating around the obstacles arriving at the goal.

From experiments, we can observe that the navigation system is capable of exploring and navigating in a previously unknown environment and reliably find its way to the global goal.
The local navigation is capable of avoiding obstacles in a reactive manner without a pre-calculated path.
Combining the motion policy with the global navigation strategy allows the robot to escape and avoid local optima.
The experimentation code with images and videos\footnote{Video: https://youtu.be/MhuhsSdzZFk} of the presented and additional experiments in static and dynamic settings are available from our repository\footnote{Repository: https://github.com/reiniscimurs/GDAE} and supplementary material.


\section{CONCLUSIONS} \label{CONCLUSIONS}

In this paper, a DRL goal-driven fully autonomous exploration system is introduced.
It is capable of arriving at a designated goal while recording the environment without direct human supervision.
As the experiments show, the system successfully combines reactive local and global navigation strategies. 
Moreover, the task of introducing a neural network-based module for an end-to-end system proves to be beneficial as it allows the robot to move without generating an explicit plan, but its shortcomings are alleviated by introducing the global navigation strategy. 
The obtained experimental results show that the proposed system works reasonably close to the optimal solution obtained by the path planner from an already known environment.
Additionally, GDAE is more reliable by relying on direct sensor inputs instead of generating plans from an uncertain map.


In the current implementation, the motion policy training was performed with the model of the same robot as in the real-life experiments.
This allowed for easy transfer of the network parameters to the embedded implementation as the network parameters were optimized for its specifications.
To generalize to various types of robots, system dynamics could be introduced as a separate input state to the neural network, and training performed accordingly.
By only providing robot dynamics, it would be possible to perform local navigation up to its best capabilities.
Additionally, a long short-term memory architecture could prove beneficial in alleviating the local optimum problem and help avoid obstacles out of the current range of sensors.
The design of such a network will be the next step of the ongoing research.

\bibliographystyle{unsrt}
\bibliography{bibliography.bib}

\end{document}